\documentclass{ifacconf}

\usepackage{graphicx}      % include this line if your document contains figures
\usepackage{epsfig}
\graphicspath{ {./images/} }
\usepackage{natbib}        % required for bibliography
\usepackage{cuted}
\usepackage{amsmath} % assumes amsmath package installed
\usepackage{amssymb}  % assumes amsmath package installed
\newcommand{\RomanNumeralCaps}[1]{\MakeUppercase{\romannumeral #1}}
\renewcommand{\thetable}{\Roman{table}}

\begin{document}
\begin{frontmatter}

\title{Learning-Based Modeling of Soft Actuators Using Euler Spiral-Inspired Curvature\thanksref{footnoteinfo}} 
% Title, preferably not more than 10 words.

\thanks[footnoteinfo]{This research was supported by the National Science Foundation (CNS 2237577, ECCS 2024649 and CMMI 1940950).}

\author[First]{Yu Mei,} 
\author[First]{Shangyuan Yuan,} 
\author[First]{Xinda Qi,} 
\author[First]{Preston Fairchild,} 
\author[First]{and Xiaobo Tan}

\address[First]{Department of Electrical and Computer Engineering, \\
Michigan State University, East Lansing, MI 48824 USA \\
(e-mail: \{meiyu1, yuanshan, qixinda, fairch42, xbtan\}@msu.edu)}

\begin{abstract}                % Abstract of not more than 250 words.
Soft robots, distinguished by their inherent compliance and continuum structures, present unique modeling challenges, especially when subjected to significant external loads such as gravity and payloads. In this study, we introduce an innovative data-driven modeling framework leveraging an Euler spiral-inspired shape representations to accurately describe the complex shapes of soft continuum actuators. Based on this representation, we develop neural network-based forward and inverse models to effectively capture the nonlinear behavior of a fiber-reinforced pneumatic bending actuator. Our forward model accurately predicts the actuator’s deformation given inputs of pressure and payload, while the inverse model reliably estimates payloads from observed actuator shapes and known pressure inputs. Comprehensive experimental validation demonstrates the effectiveness and accuracy of our proposed approach. Notably, the augmented Euler spiral-based forward model achieves low average positional prediction errors of 3.38\%, 2.19\%, and 1.93\% of the actuator length at the one-third, two-thirds, and tip positions, respectively. Furthermore, the inverse model demonstrates precision of estimating payloads with an average error as low as 0.72\% across the tested range. These results underscore the potential of our method to significantly enhance the accuracy and predictive capabilities of modeling frameworks for soft robotic systems.

\end{abstract}


\end{frontmatter}
%===============================================================================

\section{Introduction}
Intriguing properties of soft robots, such as inherent compliance and morphological adaptation, allow them to adapt to complex environments \citep{hawkes2017soft}, perform delicate tasks \citep{shintake2018soft}, and interact safely with humans \citep{polygerinos2013towards}. In recent years, considerable attention has been dedicated to the studies of soft robots, including robot design, kinematic and dynamic modeling, sensing, and control methods \citep{rus2015design, webster2010design}. While the continuum nature and infinite degrees of freedom (DOFs) enables maneuverability and flexibility for soft robots, they entail challenges in modeling, especially when the robot is subject to external forces. 

There have been a variety of modeling approaches reported for continuum robots (or soft robotic arms), which can be broadly classified into two categories: piecewise constant curvature (PCC)-based and variable curvature (VC)-based models~\citep{chen2022review}. The PCC approach has been used to model various continuum robots, such as PneuNets bending actuators (PBA)~\citep{liu2020modeling}, fiber-reinforced actuators~\citep{polygerinos2015modeling}, and soft pneumatic arms~\citep{mei2023simultaneous}. Although the PCC assumption holds approximately under moderate external forces~\citep{della2018dynamic}, it fails to capture the continuum robot shape under relatively large loads or gravity~\citep{gonthina2019modeling}, as illustrated in Fig.~1. Due to this limitation, there is an increasing trend toward developing VC-based models. For example, \cite{renda20123d} treated a soft octopus-inspired arm as a flexible, slender rod, and uses the Cosserat theory of elasticity to determine its shape in response to external forces. In contrast, \cite{wu2022fem} discretized a soft continuum robot into fine mesh elements and uses finite element method (FEM) simulation to compute their configuration. Other VC-based approaches include the absolute nodal coordinate formulation (ANCF)~\citep{huang2021kinematic} and Euler–Bernoulli beam models~\citep{fairchild2023physics}. Since these approaches typically require solving coupled partial differential equations or a large number of ordinary differential equations, the associated computational complexity and cost remain a challenge. In summary, there is a trade-off between approximation accuracy and computational efficiency among state-of-the-art kinematic models.

As a VC-based modeling approach, Euler spiral-based shape representation has recently shown strong promise ~\citep{rao2021using}. Euler spirals—also known as clothoids or Cornu spirals—were originally introduced by James Bernoulli to model elastic deformation~\citep{levien2008euler}. Defined by curvature varying linearly with the arc length, Euler spirals were first applied in robotics to describe squid-inspired shapes~\citep{gonthina2019modeling}. More recently, Euler Arc Spline models have been proposed to model tendon-driven continuum robots in both planar and 3D settings~\citep{rao2021using, rao2022shape}. These models discretize the robot into multiple constant-curvature segments with curvatures following an arithmetic progression. While efficient and reasonably accurate, they approximate the naturally smooth Euler curvature with piecewise segments.

\begin{figure}[t]
\centerline{\includegraphics[scale=0.55]{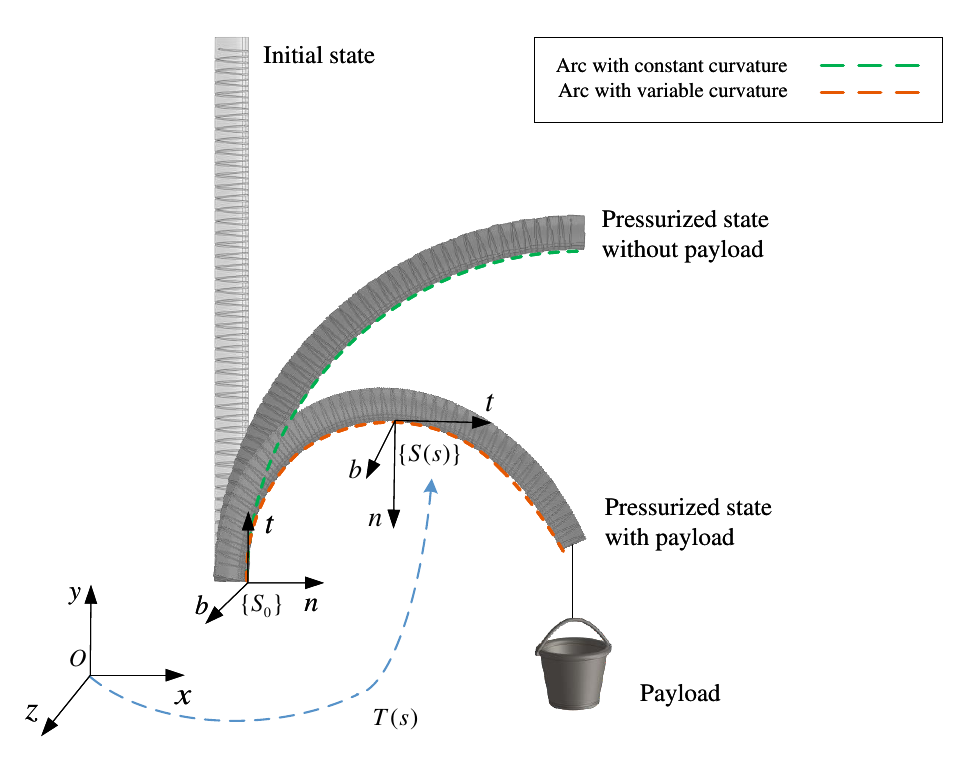}}
\caption{Schematic illustration of a soft fiber-reinforced bending actuator in different states. The bending actuator is straight when there is no air pressure initially, and bends with a constant curvature when the air inflates the inner chamber in the absence of loading. Under a payload applied on the tip, the actuator deforms into a shape with a continuously varying curvature, which could be approximated by an Euler spiral-inspired representation.}
\label{fig1}
\end{figure}

In this paper, we propose a systematic data-driven approach for modeling soft continuum robots with \textit{continuously varying} curvatures, induced by both actuation and external forces, using an Euler spiral-inspired representation. In this formulation, the curvature profile is modeled as a polynomial function of arc length, enabling compact and flexible shape parameterization. The validity of this representation is supported by the comparison with a physical model based on Euler–Bernoulli beam theory. To learn the mapping between actuation inputs, external forces, and robot configurations, we develop data-driven forward and inverse models based on multilayer perceptrons (MLPs). To construct the training dataset, we introduce an efficient method for extracting curvilinear parameters by solving a \(G^1\) Hermite interpolation problem, using the positions and orientations of the robot’s proximal and distal ends.

The remainder of the paper is organized as follows. Section \RomanNumeralCaps{2} introduces the Euler spiral-inspired representation, parameter identification method, and MLP architecture. Section \RomanNumeralCaps{3} presents the validation and dataset collection processes. Section \RomanNumeralCaps{4} reports experimental results for model training and evaluation. Conclusions and future directions are discussed in Section \RomanNumeralCaps{5}.

\section{METHODS}
In this section, the fundamentals of Euler spiral are introduced, and a Euler spiral-inspired shape representation is proposed for fiber-reinforced soft actuators. We then formulate a ${G^1}$ Hermite interpolation problem for acquiring the curvilinear parameters for shape representation, given the configurations (position and orientation) of the actuator at both ends. Finally, we describe the neural networks for the forward and inverse models for the soft actuator under external loading.

\vspace{-0.4em}
\subsection{Shape Representation inspired by Euler Spiral}
A general planar Euler spiral satisfies the following equations \citep{bertolazzi2015g1}:
\begin{equation} \label{eq:1}
\left\{ {\begin{array}{*{20}{c}}
  {\dot x(s) = \cos \theta (s),}&{x(0) = {x_0},} \\ 
  {\dot y(s) = \sin \theta (s),}&{y(0) = {y_0},} \\ 
  {\dot \theta (s) = \kappa (s),}&{\theta (0) = {\theta _0},} 
\end{array}} \right.
\end{equation}
where $s$ is the curvilinear abscissa also known as the arc length parameter, $(x(s),y(s))$ are the coordinates of the point $s$, and $\theta(s)$ is the orientation of the curve at $s$, which is also the direction of the tangent $(\dot x(s), \dot y(s))$, as depicted in Fig. 2.
\begin{figure}[b]
\centerline{\includegraphics[scale=0.55]{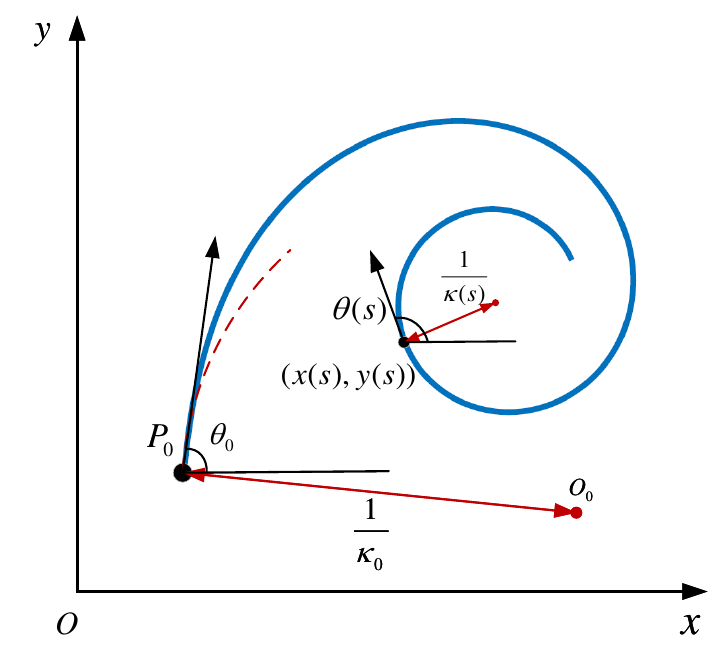}}
\caption{Illustration of the Euler spiral.}
\label{fig2}
\end{figure}
Here $(x(0),y(0)),\theta(0)$ are the position and the curve orientation angle at the base point $P_0$, and $\kappa(s)$ is the curvature at point $s$. Inspired by the geometric properties of the Euler spiral, the curvature $\kappa(s)$ along the arc length $s$ is expressed as a polynomial of order $N$: 
\begin{equation} \label{eq:2} 
\kappa(s) = \sum_{k=0}^N \kappa_k s^k, 
\end{equation} 
where $\kappa_k$ denotes the coefficient corresponding to the $k$-th order term. Notably, the Euler spiral corresponds to the case of $N = 1$, a constant curvature segment is captured by a single nonzero term $\kappa_0$, and a straight line is represented when all coefficients $\kappa_k$ are zero. By plugging Eq. (\ref{eq:2}) into Eq. (\ref{eq:1}), one can get: 
\begin{equation} \label{eq:3}
\left\{ \begin{gathered}
  x(s) = \int_0^s {\cos (\theta (\tau ))} d\tau + {x_0} , \hfill \\
  y(s) = \int_0^s {\sin (\theta (\tau ))} d\tau + {y_0} , \hfill \\
  \theta (s) = \sum\limits_{k = 0}^N {\frac{{{\kappa _k}}}{{k + 1}}} {s^{k + 1}} + {\theta _0}, \hfill \\ 
\end{gathered}  \right.
\end{equation}
which implies that the entire curve is fully characterized by the curvature coefficients $\{ \kappa _k\}$ and initial condition $ (x_0,y_0)$ and $\theta_0$.

In this work, we focus on the shape reconstruction of a single-segment pneumatic bending actuator under tip loading. As shown in Fig. 1, a Frenet-Serret frame $\{S(s)\}$ is introduced to describe the orientation of the curve at each point of the bottom layer of the bending actuator \citep{webster2010design}, where the vector ${\bf{t}}$ is always tangent to the curve, the vector $\bf{n}$ is normal to the curve, and the vector $\bf{b}$ is perpendicular to the bending plane. By considering $x$-$y$ as the bending plane and aligning the origin, one can obtain the homogeneous transformation $T(s)$ from the inertial frame to the frame $\{S(s)\}$ using parameterized Euler curves:
\begin{equation} \label{eq:4}
\left\{ \begin{array}{l}
T(s) = \left[ {\begin{array}{*{20}{c}}
{R(s)}&{p(s)}\\
0&1
\end{array}} \right],\\
p(s) = {[\begin{array}{*{20}{c}}
{x(s)}&{y(s)}&0
\end{array}]^T},\\
R(s) = {R_z}(\theta (s)),
\end{array} \right.
\end{equation}
where $x(s), y(s), \theta(s)$ are as defined in Eq. (\ref{eq:3}) and ${R_z}(\theta (s)) \in SO(3)$ indicates a rotation about the $+z$-axis by the angle $\theta (s)$. Note that the arc length parameter $s \in [0, L]$. As a result, with the given curvilinear parameter $\boldsymbol{\kappa} = [\kappa_0, \kappa_1, \dots, \kappa_N]^T$
 and the initial conditions $(x_0, y_0), \theta_0$, the actuator shape can be derived through Eq. (\ref{eq:3}) and Eq. (\ref{eq:4}). Since the base point is fixed, the initial coordinates $(x_0, y_0)$ are fixed. Consequently, the shape representation parameters of the fiber-reinforced actuator are is given by: ${\mathbf{q}} = {[\boldsymbol{\kappa}^T,{\theta _0}]^T}$.
 
\subsection{${G^1}$ Hermite Interpolation to Obtain Curvilinear \\Parameters}
As discussed above, the shape representation $\bf{q}$ is needed to construct the actuator's shape. However, it is practically difficult to find $\bf{q}$ since the direct and accurate measurement of the curvature is not readily available. We propose an efficient approach to the extraction of the shape representation by formulating a ${G^1}$ Hermite interpolation problem, which is defined as finding an interpolation curve that matches the tangent vectors at two end points \citep{bertolazzi2015g1, zhou2012euler}. Since tangent vectors at the end points can be easily captured by a vision system, we can leverage the property of spiral to solve the interpolation problem. As illustrated in Fig. 3, we aim to find the optimal curvilinear parameters ${\kappa}^*$ to build a spiral curve that solves the ${G^1}$ Hermite interpolation problem with two given points $P_0, P_1$ and the corresponding orientation angles, ${\theta _0}$ and ${\theta _1}$. In particular, this interpolation problem is formulated as the following optimization problem subject to the given boundary conditions:
\begin{equation} \label{eq:opt_full}
\begin{aligned}
\boldsymbol{\kappa}^* 
&= \operatornamewithlimits{arg\,min}_{\boldsymbol{\kappa} \in \mathbb{R}^{N+1}} 
\left\|
\begin{bmatrix}
x(L; \boldsymbol{\kappa}) - x_1 \\
y(L; \boldsymbol{\kappa}) - y_1 \\
\theta(L; \boldsymbol{\kappa}) - \theta_1
\end{bmatrix}
\right\|_2^2 \\
\text{s.t.} \quad 
& x(0) = x_0,\quad y(0) = y_0, \\
& \dot{x}(0) = \cos\theta_0,\quad \dot{y}(0) = \sin\theta_0, \\
& x(L) = x_1,\quad y(L) = y_1, \\
& \dot{x}(L) = \cos\theta_1,\quad \dot{y}(L) = \sin\theta_1.
\end{aligned}
\end{equation}
Here, \(x(L; \boldsymbol{\kappa})\) and \(y(L; \boldsymbol{\kappa})\) represent the terminal position of the curve at arc length \(L\), parameterized by the curvilinear parameters $\boldsymbol{\kappa}$. These quantities are obtained by integrating via Eq.~(3).

\begin{figure}[h]
\centerline{\includegraphics[scale=0.45]{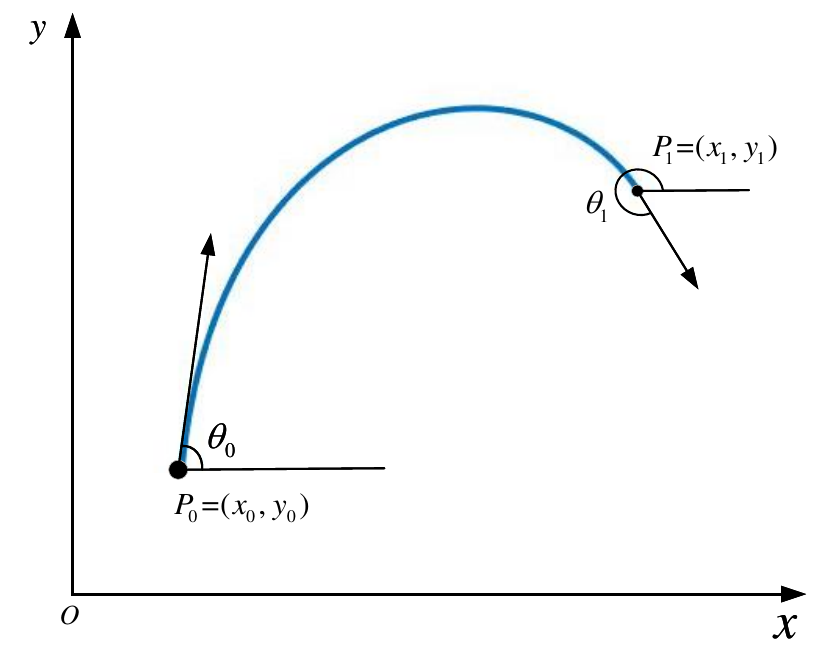}}
\caption{${G^1}$ Hermite interpolation schema and notation.}
\label{fig3}
\end{figure}

\subsection{Neural Network-based Forward and Inverse Models}
The forward model of the soft actuator aims to predict the shape of the actuator under given inflation pressure $P$ and the tip payload $W$. On the other hand, the inverse model aims to infer the weight $W$ of payload based the shape of the actuator and the inflation pressure $P$. It has been shown that neural networks are an effective tool for learning the kinematics of soft robots, which are highly nonlinear \citep{truby2020distributed, grassmann2022dataset}. In this work, MLPs are designed to represent the forward and inverse models, which are trained separately. In particular, for the forward model, the input is ${[P,W]^T}$ and the output is shape representation vector $\bf{q}$. For the inversion model, $[\boldsymbol{\kappa}^T, \theta_0, P]^T$ and $W$ serve as the input and the output, respectively. 

\section{Numerical Validation}

The effectiveness of using a spiral curve to represent the shape of a bending actuator under external loading is supported by physics-based simulation results based on Euler--Bernoulli beam theory. Due to space limitations, the detailed modeling is not included here, but a comprehensive description is available in our prior work~\citep{mei2024simultaneous}. In principle, the dimension $N$ of the curvilinear parameter can be arbitrarily large to accurately capture complex shapes. However, to ensure computational efficiency and suitability for 2D planar bending actuators, we aim to identify a relatively low-order polynomial that still achieves adequate shape representation. In this work, we examine both the original Euler spiral formulation with $N=1$ and an augmented formulation with $N=2$.

We simulated the actuator shapes under varying actuation pressures and tip loads using Euler--Bernoulli beam theory. Two of the most common types of tip loads relevant to practical applications of bending actuators—payload and contact forces—are considered, as illustrated in Fig.~4.

\begin{figure}[h]
\centerline{\includegraphics[scale=0.55]{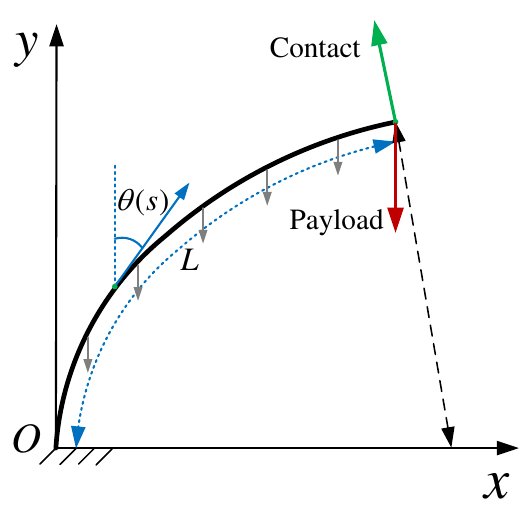}}
\caption{Illustration of soft actuator under tip loads.}
\label{fig4}
\end{figure}

\begin{figure}[h]
\centerline{\includegraphics[scale=0.45]{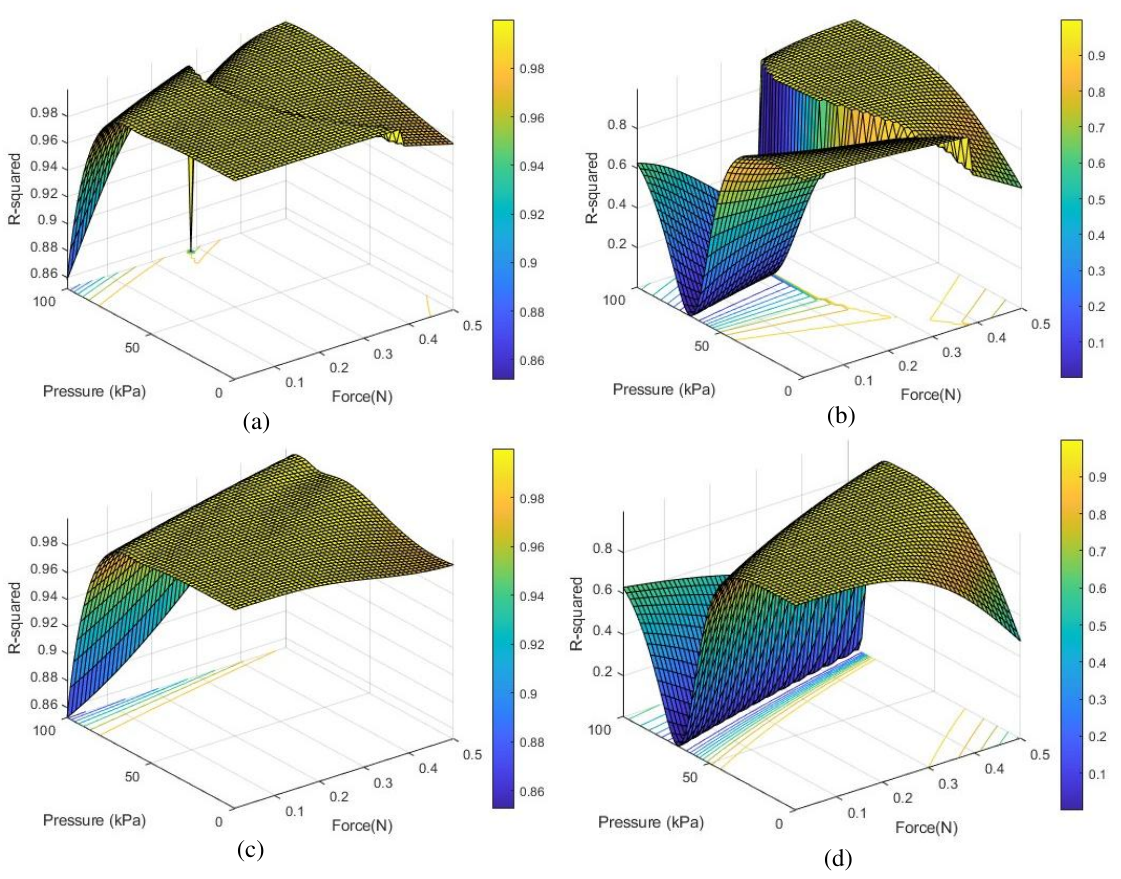}}
\caption{Regression results using linear and quadratic functions for different loading scenarios. (a–b) Quadratic and linear fitting under payload conditions. (c–d) Quadratic and linear fitting under contact conditions.}
\label{fig5}
\end{figure}

By varying the pressure from 0 to 100 kPa and the payload from 0.1 to 0.5 N, we simulated the curvature distribution along the arc length for each data point using the physical modeling. The simulated curvature profiles were then fit using linear and quadratic functions, corresponding to \(N = 1\) and \(N = 2\), respectively. The regression results are shown in Fig.~\ref{fig5}, where higher \(R^2\) values indicate better fitting performance. The results show that the quadratic curvature model achieves \(R^2 = 0.99 \pm 0.01\) and \(0.99 \pm 0.02\) for the payload and contact cases, respectively, compared to \(0.77 \pm 0.31\) and \(0.76 \pm 0.33\) for the linear curvature model. These results indicate that the proposed spiral curve formulation with \(N = 2\) provides a more accurate approximation of the actuator shape.

\section{Experimental Setup and Dataset Construction}
An experimental platform was developed to collect data for model learning and validation, as shown in Fig.~6. The platform consists of a soft fiber-reinforced bending actuator~\citep{polygerinos2015modeling} and a vision-based detection system using colored markers, following the same setup described in~\citep{mei2024simultaneous}. The actuator has a length of 150 mm and weighs 25.86 g.

\begin{figure}[h]
\centerline{\includegraphics[scale=0.45]{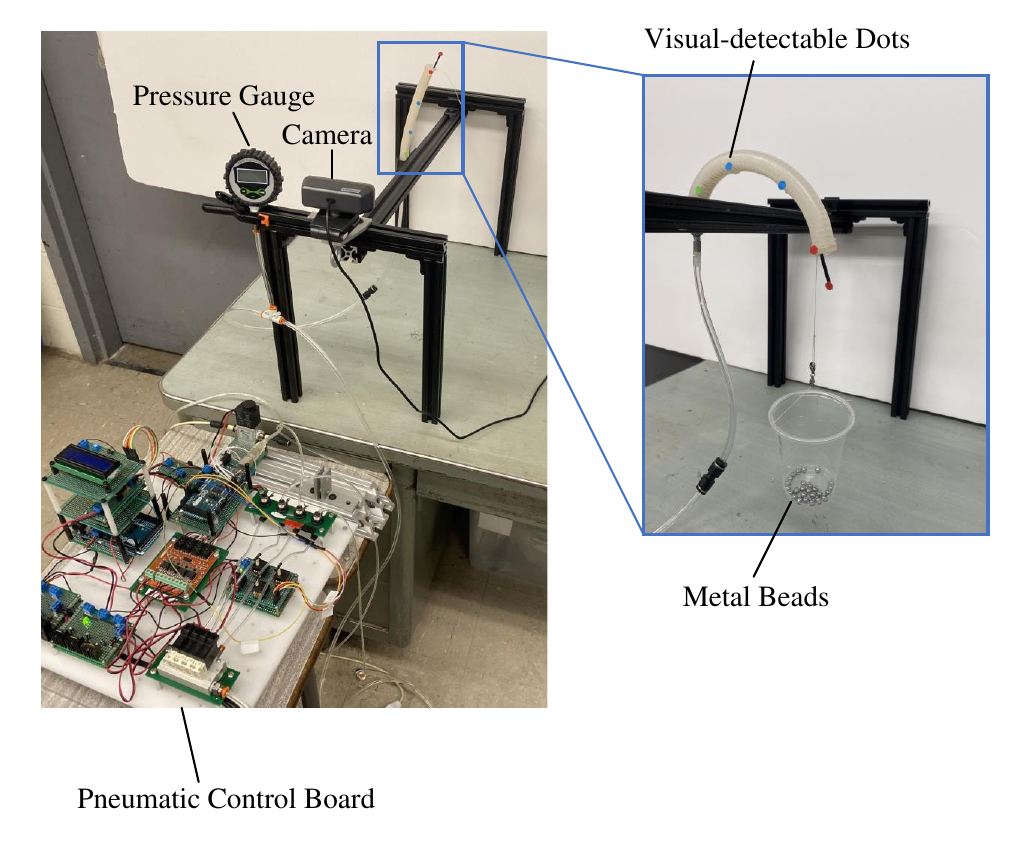}}
\caption{Experimental setup for dataset collection.}
\label{fig6}
\end{figure}

\begin{figure*}[t]
    \centerline{\includegraphics[scale=0.48]{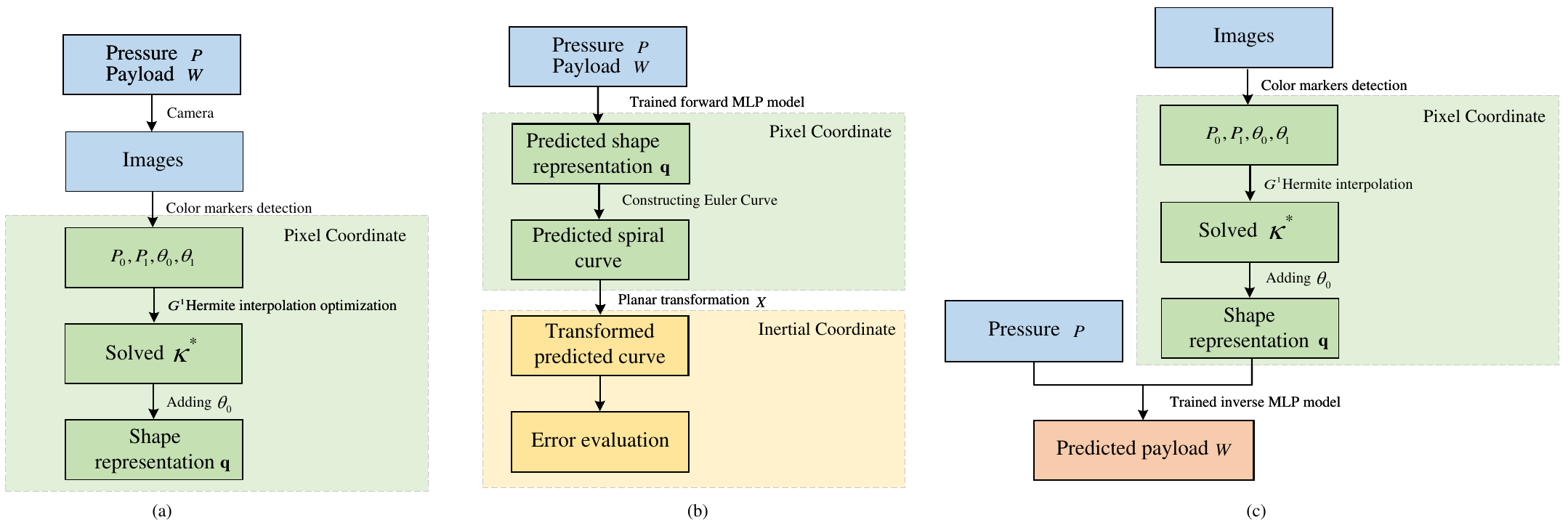}}
    \caption{The flowchart of (a) collecting dataset, (b) predicting the shape via the trained forward MLP model, (c) predicting the payload via trained the inverse MLP model. }
    \label{fig7}
\end{figure*}

With the above experimental setup, shape data of the soft bending actuator can be collected and automatically labeled using the computer vision system. In this work, we collected the dataset under payload conditions only, without loss of generality. Fig. 7 (a) shows the process of constructing the dataset: images are first captured by the camera when the actuator reaches the steady state under given inflation pressure $P$ and payload $W$, which are then used to extract automatically the position of the end point $P_1$, the orientation angles at the base and at the distal end, $\theta_0$ and $\theta_1$. The ${G^1}$ Hermite interpolation optimization in Eq.~(5) is then employed to computes the curvilinear parameters $\kappa$. With added $\theta_0$, the shape representation $\bf{q}$ in the pixel coordinates can be obtained. As for various input data, inflation pressure is varied from 20 kPa to 100 kPa, with steps of 10 kPa, and the tip loading is varied from 3.61 g to 29.05 g with steps of 0.254 g. The variable payload is realized by placing a total of 100 metal beads into the cup hung on the tip, one after another. Each metal bead has a diameter of 4 mm and weights 0.254 g, and the cup with hook weights 3.61 g. Note that, the range of pressure is chosen to avoid the buckling phenomenon of the actuator under low pressures and potential damage to the actuator under high pressures. The range of the weights is determined so that the markers are visible. Finally, 900 measured data points are obtained, where each data point consists of ${P,W,{\theta _0},{\kappa}}$, and the positions of five color markers in the pixel coordinates.

\section{EXPERIMENTAL RESULTS}
\subsection{Training of MLPs}
\vspace{-2pt}
As discussed in Section \RomanNumeralCaps{2}-C, two MLPs are trained separately for the forward and inverse models, implemented in \texttt{TensorFlow}. Each layer is a fully connected neural network layer, initialized using \texttt{He-uniform} for the weight matrix and employing \texttt{ReLU} as the nonlinear activation function. To reduce the impact of pneumatic chattering noise, the input vector is smoothed using a Gaussian-weighted moving average filter.

The forward model is implemented as a three-layer MLP with 64, 32, and 16 neurons, while the inverse model uses two layers with 16 and 8 neurons. Both networks are trained for 200 epochs using the \texttt{Adam} optimizer (learning rate 0.001, decay \(5 \times 10^{-6}\)) to minimize the mean absolute error (MAE) loss.

\subsection{Model Prediction Performance}
\textit{1) Forward model:} With the trained forward model, the shape of the soft actuator can be predicted in the pixel coordinates from the given actuation pressure $P$ and payload $W$, as illustrated in Fig. 7 (b). Then the predicted spiral curve in the pixel coordinates is converted into the task space with the calibrated pixel-real mapping. Error evaluation is performed at three reference points located at the one-third, two-thirds, and the end of the actuator length, as identified by the two blue markers and one red marker. The error is computed as the Euclidean distance error normalized by the length of the actuator. We compared three different curvilinear parameters $\kappa$ with $N=0,1,2$ for shape representation, where $N=0$ is the constant curvature, $N=1$ is the classical Euler spiral and $N=2$ is augmented Euler spiral.

We randomly picked 40 data points for validation, which were not used during model training. The average task-space errors and corresponding standard deviations are reported in Table~\ref{tab:table1}. To further illustrate the results visually, we highlight two examples with \([P, W]^T = [40,\, 23.7]^T\) and \([P, W]^T = [100,\, 9.0]^T\), and plot the predicted shapes using the proposed spiral curve models, as shown in Fig.~8. From both Table~\RomanNumeralCaps{1} and Fig.~8, it is evident that the Euler spiral and augmented Euler spiral approaches outperform the constant-curvature model. Moreover, for data points within the mid-range of the workspace, predictions from both \(N=1\) and \(N=2\) spiral models are comparable, as shown in Fig.~8(a). However, under large deformations resulting from high pressure or payload, the augmented Euler spiral (\(N=2\)) provides improved accuracy, as shown in Fig.~8(b).

\begin{table}[h]
\centering
\footnotesize
\caption{Average errors (\%) at three reference points relative to actuator length.}
\label{tab:table1}
\begin{tabular}{c p{2.2cm} p{2.2cm} p{2.2cm}}
\hline
 & 1/3 position error (\%) & 2/3 position error (\%) & Tip error (\%) \\ \hline
N=2 & 3.38 $\pm$ 0.21 & 2.19 $\pm$ 0.40 & 1.93 $\pm$ 1.33 \\
N=1 & 3.43 $\pm$ 0.18 & 2.50 $\pm$ 0.45 & 2.02 $\pm$ 1.58 \\
N=0 & 6.67 $\pm$ 4.24 & 17.5 $\pm$ 19.0 & 35.0 $\pm$ 35.0 \\
\hline
\end{tabular}
\end{table}

\begin{figure}[h]
\centerline{\includegraphics[scale=0.4]{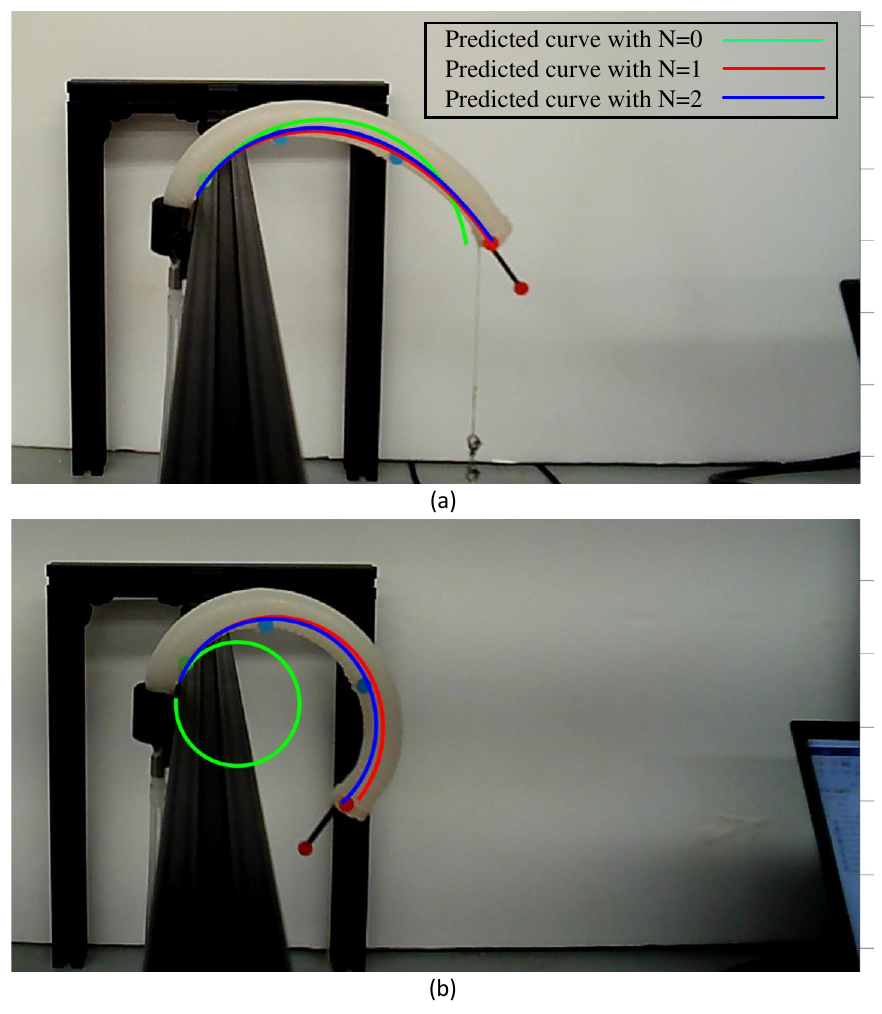}}
\vspace{-5pt}
\caption{Comparisons of the predictions from the proposed spiral curve with different curvilinear parameters. (a) results under 40 kPa pressure and 23.7 g payload. (b) results under 100 kPa pressure and 9.0 g payload.}
\label{fig8}
\end{figure}

\vspace{-3pt}
\begin{table}[b]
\centering
\footnotesize
\caption{Average tip errors (\%) with respect to the range of the payload for different orders.}
\label{tab:table_tip}
\vspace{-3pt}
\begin{tabular}{lc}
\hline
\textbf{Order} & \textbf{Load Error (\%)} \\
\hline
N = 2 & 0.72 $\pm$ 0.62 \\
N = 1 & 1.08 $\pm$ 1.12 \\
N = 0 & 1.21 $\pm$ 1.08 \\
\hline
\end{tabular}
\end{table}

\textit{2) Inverse model:} Fig. 7 (c) shows how the trained inverse model is used to infer the weight of the payload. The inputs are the image from the camera and the air pressure detected from the gauge. The position and curvature orientation at the initial point and the end point ${\left[ {{P_0},{P_1},{\theta _0},{\theta _1}} \right]}$ are computed based on detected color markers. After solving ${G^1}$ Hermite interpolation optimization, the curvilinear parameter $\kappa^*$ can be obtained and the corresponding shape representation $\bf{q}$ can also be extracted. Finally, after we feed the shape vector $\bf{q}$ and the input pressure $P$ into the trained inverse model, the unknown payload can be predicted. We also validate the inverse model using the same data points in the validation of the forward model that were not used in the model training, as shown in Table~\RomanNumeralCaps{2}. The results show that the prediction of the inverse model is accurate with respect to the payload range (25.44 g), considering that the resolution of the payload is 1 \%.

\section{Conclusion and Future work}
In this work, we proposed an accurate and computationally efficient data-driven modeling approach for soft continuum robots, inspired by the Euler spiral. Leveraging its variable-curvature formulation, we introduced a compact shape representation for robots under external forces. Using \(G^1\) Hermite interpolation, we developed a framework to automatically extract the curvilinear parameters. Neural network models were then trained to predict robot shapes under actuation and loading, and to estimate external forces from observed shapes. The effectiveness of the approach was validated experimentally.

While this study focused on a single soft bending actuator, future work will extend the framework to multi-segment continuum arms and explore more general 3D deformations and external forces. We also aim to incorporate dynamics into the model and investigate its integration with learning-based controllers for soft robots.

\begin{ack}
We gratefully acknowledge Xinyu Zhou for his discussion.  
\end{ack}

\bibliography{ifacconf}             % bib file to produce the bibliography
                                                     % with bibtex (preferred)
                                                   
%\begin{thebibliography}{xx}  % you can also add the bibliography by hand

%\bibitem[Able(1956)]{Abl:56}
%B.C. Able.
%\newblock Nucleic acid content of microscope.
%\newblock \emph{Nature}, 135:\penalty0 7--9, 1956.

%\bibitem[Able et~al.(1954)Able, Tagg, and Rush]{AbTaRu:54}
%B.C. Able, R.A. Tagg, and M.~Rush.
%\newblock Enzyme-catalyzed cellular transanimations.
%\newblock In A.F. Round, editor, \emph{Advances in Enzymology}, volume~2, pages
%  125--247. Academic Press, New York, 3rd edition, 1954.

%\bibitem[Keohane(1958)]{Keo:58}
%R.~Keohane.
%\newblock \emph{Power and Interdependence: World Politics in Transitions}.
%\newblock Little, Brown \& Co., Boston, 1958.

%\bibitem[Powers(1985)]{Pow:85}
%T.~Powers.
%\newblock Is there a way out?
%\newblock \emph{Harpers}, pages 35--47, June 1985.

%\bibitem[Soukhanov(1992)]{Heritage:92}
%A.~H. Soukhanov, editor.
%\newblock \emph{{The American Heritage. Dictionary of the American Language}}.
%\newblock Houghton Mifflin Company, 1992.

%\end{thebibliography}
\end{document}